\documentclass{article}

\usepackage{arxiv}

\usepackage[utf8]{inputenc} 
\usepackage[T1]{fontenc}    
\usepackage{hyperref}       
\usepackage{url}            
\usepackage{booktabs}       
\usepackage{amsfonts}       
\usepackage{nicefrac}       
\usepackage{microtype}      
\usepackage{lipsum}		
\usepackage{graphicx}
\usepackage{natbib}
\usepackage{doi}
\usepackage{amsmath}

\usepackage{xcolor}  
\usepackage{authblk}
\usepackage{wrapfig}
\usepackage{multirow}
\usepackage{subcaption}

\title{Fast Quantum Property Prediction via Deeper 2D and 3D Graph Networks}

\author[*, 1]{\textbf{Meng Liu}}
\author[*, 1]{\textbf{Cong Fu}}
\author[1]{\textbf{Xuan Zhang}}
\author[1]{\textbf{Limei Wang}}
\author[1]{\textbf{Yaochen Xie}}
\author[1]{\textbf{Hao Yuan}}
\author[1]{\textbf{Youzhi Luo}}
\author[1]{\textbf{Zhao Xu}}
\author[2]{\textbf{Shenglong Xu}}
\author[1]{\textbf{Shuiwang Ji}}
\affil[*]{These authors contributed equally to this work.}
\affil[1]{Department of Computer Science and Engineering, Texas A\&M University, College Station, TX 77843, USA}
\affil[2]{Department of Physics and Astronomy, Texas A\&M University, College Station, TX 77843, USA}
\affil[ ]{\texttt{\{mengliu,congfu,xuan.zhang,limei,ethanycx,hao.yuan,yzluo,zhaoxu,slxu,sji\}@tamu.edu}}

\date{}

\renewcommand{\undertitle}{Technical Report}

\hypersetup{
pdftitle={A template for the arxiv style},
pdfsubject={q-bio.NC, q-bio.QM},
pdfauthor={David S.~Hippocampus, Elias D.~Striatum},
pdfkeywords={First keyword, Second keyword, More},
}

\begin{document}
\maketitle

\begin{abstract}
	Molecular property prediction is gaining increasing attention due to its diverse applications. One task of particular interests and importance is to predict quantum chemical properties without 3D equilibrium structures. This is practically favorable since obtaining 3D equilibrium structures requires extremely expensive calculations. In this work, we design a deep graph neural network to predict quantum properties by directly learning from 2D molecular graphs. In addition, we propose a 3D graph neural network to learn from low-cost conformer sets, which can be obtained with open-source tools using an affordable budget. We employ our methods to participate in the 2021 KDD Cup on OGB Large-Scale Challenge (OGB-LSC), which aims to predict the HOMO-LUMO energy gap of molecules. Final evaluation results reveal that we are one of the winners with a mean absolute error of 0.1235 on the holdout test set. Our implementation is available as part of the MoleculeX package (\url{https://github.com/divelab/MoleculeX}).
\end{abstract}


\section{Introduction}


Molecular property prediction is of great importance in many applications, such as chemistry, drug discovery, and material science. Many molecular properties, such as the energy and the shape of molecules, could be computed by quantum mechanical simulation methods, such as Density Functional Theory (DFT). However, such methods are computationally expensive. To be specific, it takes around several hours to run the DFT 
computation for a small molecule since it requires geometry optimization to obtain 3D equilibrium structures. Therefore, it is highly desired if we could predict such quantum chemical properties without 3D equilibrium structures of molecules.

 Recently, graph deep learning methods have been developed for molecular property prediction~\citep{gilmer2017neural,wu2018moleculenet,yang2019analyzing,stokes2020deep,wang2020advanced}. As molecules can be naturally treated as graphs by viewing atoms as nodes and bonds as edges, these methods leverage various graph neural networks (GNNs)~\citep{bronstein2017geometric,gilmer2017neural,battaglia2018relational,kipf2016semi,velivckovic2017graph,xu2018powerful} to learn from 2D molecular graphs and achieve great success. Nevertheless, these approaches only employ shallow graph neural networks, thus limiting the expressive power and the receptive fields of the models. In addition, the existing methods only focus on 2D molecular graphs without explicitly considering 3D information, which is crucial for determining quantum chemical properties. Hence, it remains challenging to incorporate useful 3D information with an affordable budget since 3D equilibrium structures are usually unavailable and time-consuming to obtain. While there is another line of research that predicts quantum chemical properties given 3D equilibrium structures~\citep{schutt2018schnet,klicpera2019directional,liu2021spherical,hu2021forcenet}, it is orthogonal to our work since 3D equilibrium structures are not available in this work.

Motivated by the above two challenges, we propose a deeper 2D graph neural network on 2D molecular graphs and a 3D graph neural network on low-cost conformer sets to predict quantum chemical properties. Specifically, we leverage the recent advanced deep graph neural networks, namely DeeperGCN~\citep{li2020deepergcn} and DAGNN~\citep{liu2020towards}, to construct our 2D model. For the 3D model, we obtain the conformer sets using RDKit~\citep{landrum2006rdkit}, which takes averagely less than $0.05$ seconds for a molecule. Afterwards, we deploy a 3D GNN to learn from the conformer sets. Our intuition is to utilize the imprecise but descriptive 3D information contained in the conformer sets to improve the prediction performance for quantum chemical properties. To show the effectiveness of our proposed approach, we conduct experiments on PCQM4M-LSC dataset~\citep{hu2021ogb}, inlcluded in the 2021 KDD Cup on OGB Large-Scale Challenge\footnote{\url{https://ogb.stanford.edu/kddcup2021/}}, to predict the HOMO-LUMO energy gap of molecules without given 3D equilibrium structures. The results show that our methods achieve remarkable prediction performance.

\section{Methodology}
We consider a molecular graph $\mathcal{G}=(\mathcal{V}, \mathcal{E}, \mathbf{X}, \mathbf{E}, \mathbf{A})$, where node set $\mathcal{V}=\{1,2,\cdots,n\}$ denotes atoms and edges $\mathcal{E} \subseteq \mathcal{V}\times\mathcal{V}$ are given by bonds or by connecting atoms with considering certain cutoff distance. Without losing generality, we consider each node and edge is associated with a feature vector. To be specific, $\mathbf{X} \in \mathbb{R}^{|\mathcal{V}| \times d}$ is the node feature matrix, where each row $\mathbf{x}_i \in \mathbb{R}^d$ represents the $d$-dimensional feature vector of atom $i$. $\mathbf{E} \in \mathbb{R}^{|\mathcal{E}| \times p}$ denotes the edge feature matrix and each row is a $p$-dimensional feature vector of an edge. We use $\mathbf{e}_{ij} \in \mathbb{R}^p$ to denote the feature vector of the edge from atom $i$ to $j$. The connectivity of the graph is described by the adjacency matrix $\mathbf{A} \in \mathbb{R}^{n \times n}$.

\subsection{Deeper 2D Model on 2D Molecular Graphs}
\label{sec:2d}

We aim to construct a deep graph neural network, which can bring us large receptive fields and powerful expressivity. Our model is developed based on the recent progress on deep GNNs, particularly, DeeperGCN~\citep{li2020deepergcn} and DAGNN~\citep{liu2020towards}. The details of the main components are described below.

\paragraph{GENConv.} Most modern GNNs follow a message passing scheme~\citep{gilmer2017neural,battaglia2018relational}. To be specific, we iteratively update the node representation by aggregating information, \emph{a.k.a.}, message, from neighboring nodes. GENConv~\citep{li2020deepergcn} also follows such a mechanism. The key characteristic of GENConv is the \textit{SoftMax\_Agg} aggregation function. The forward computation of one layer GENConv for each node $i$ is formulated as
\begin{equation}
    \begin{aligned}
    &\mathbf{m}_{ji} = \textrm{ReLU}(\mathbf{x}_j+\mathbf{e}_{ji})+\epsilon,\  j\in \mathcal{N}_i \\
    &\mathbf{x}'_i = \textrm{MLP}\left(\mathbf{x}_i+\textit{SoftMax\_Agg}_\beta\left(\{\mathbf{m}_{ji}:j\in \mathcal{N}_i\}\right)\right) \\
    &\textit{SoftMax\_Agg}_\beta(\cdot) = \sum_{j\in \mathcal{N}_i}\frac{\exp{(\beta\mathbf{m}_{ji})}}{\sum_{k\in \mathcal{N}_i}\exp{(\beta\mathbf{m}_{ki})}}\cdot\mathbf{m}_{ji},
    \end{aligned}
\end{equation}
where $\mathbf{m}_{ji}$ represents the message from atom $j$ to atom $i$. $\epsilon$ is a small positive constant. $\textrm{ReLU}(\cdot)$~\citep{krizhevsky2012imagenet} is the activation function and $\textrm{MLP}(\cdot)$ denotes a multi-layer perceptron. Note that $\beta$ is a learnable scalar parameter and initialized as $1$. As shown in the above equations, in GENConv, we firstly construct messages from neighboring nodes and then aggregate them to obtain the updated representation $\mathbf{x}'_i$ for each node $i$.

\paragraph{DeeperGCN Layer.} Based on GENConv block, DeeperGCN Layer~\citep{li2020deepergcn} is proposed by further introducing skip connections~\citep{he2016deep} and the pre-activation technique~\citep{he2016identity}. Therefore, the resulting DeeperGCN Layer consists of the following components: Normalization $\rightarrow$ Activation $\rightarrow$ GENConv $\rightarrow$ Addition. We apply BatchNorm (BN)~\citep{ioffe2015batch} as the normalization and ReLU as the activation function.

\paragraph{Virtual Node.} Virtual node is found to be effective across various graph-level tasks~\citep{gilmer2017neural,hu2020open}. Generally, we can augment a graph with a virtual node that communicates with all other nodes, thus better capturing the global or long-range information. The feature of the virtual node is denoted as $\mathbf{g}$ and initialized as a zero-values vector. In each layer, the virtual node feature is updated as
\begin{equation}
\label{eq:virtual}
    \mathbf{g}' = \textrm{MLP}\left(\textrm{Readout}_\textrm{sum}\left(\{\mathbf{x}_i, i \in \mathcal{V}\}\right) + \mathbf{g}\right),
\end{equation}
where $\textrm{Readout}_\textrm{sum}(\cdot)$ represents the summation readout function.


\paragraph{DAGNN.} We apply $L$ DeeperGCN Layers and incorporate the virtual node technique to learn the node representations. We denote the resulting node representation matrix as $\mathbf{Z} \in \mathbb{R}^{|\mathcal{V}| \times f}$, where $f$ is the number of hidden dimensions. As shown by~\citet{li2021neuralfingerprint}, the performance could be further improved if we deploy a diffusion algorithm on the resulting node representations $\mathbf{Z}$. In our work, we choose DAGNN~\citep{liu2020towards}, which is demonstrated to be effective and can adaptively balance the information from various receptive fields for each node. In the original DAGNN, there are three main steps, including transformation, propagation, and adaptive adjustment. An MLP is used for transformation in the orginal DAGNN. In this work, we remove this transformation step since we can regard the previous DeeperGCN Layers as the transformation. Hence, we deploy $K$ steps of propagation and the adaptive adjustment mechanism to integrate the information of different receptive fields. Formally, the process can be represented as
\begin{equation}
    \begin{aligned}
    & \mathbf{H}^{(k)}=\mathbf{\hat{A}}^k\mathbf{Z}, \ k=1, 2, \cdots, K  & & \in \mathbb{R}^{|\mathcal{V}| \times f}\\
    & \mathbf{\tilde{H}} = \textrm{Stack}\left(\mathbf{Z}, \mathbf{H}^{(1)}, \cdots, \mathbf{H}^{(K)}\right) & & \in \mathbb{R}^{|\mathcal{V}| \times (K+1) \times f}\\
    & \mathbf{S} = \sigma\left(\mathbf{\tilde{H}}\mathbf{s}\right) & & \in \mathbb{R}^{|\mathcal{V}| \times (K+1) \times 1} \\
    & \mathbf{\tilde{S}} = \textrm{Reshape}\left(\mathbf{S}\right) & & \in \mathbb{R}^{|\mathcal{V}| \times 1 \times (K+1)} \\
    & \mathbf{Y} = \textrm{Squeeze}\left(\mathbf{\tilde{S}}\mathbf{\tilde{H}}\right) & & \in \mathbb{R}^{|\mathcal{V}| \times f},
    \end{aligned}
\end{equation}
where $\mathbf{\hat{A}}=\mathbf{\tilde{D}}^{-\frac{1}{2}}\mathbf{\tilde{A}}\mathbf{\tilde{D}}^{-\frac{1}{2}}$ is the normalized adjacency matrix, $\mathbf{\tilde{A}}=\mathbf{A}+\mathbf{I}$ is the adjacency matrix with self-loops, and $\mathbf{\tilde{D}}$ is the degree matrix corresponding to $\mathbf{\tilde{A}}$. In addition, $\mathbf{s} \in \mathbb{R}^{f\times 1}$ is a learnable vector,  $\sigma(\cdot)$ denotes the $\textrm{Sigmoid}$ function, and $\mathbf{Y}$ denotes the obtained node representation matrix. Intuitively, DAGNN utilizes a learnable vector $\mathbf{s}$ shared by all nodes to obtain retainment scores for different receptive fields, which help adaptively integrate the information of different receptive fields for each node.

\begin{figure*}[t]
	\centering
	\includegraphics[width=\textwidth]{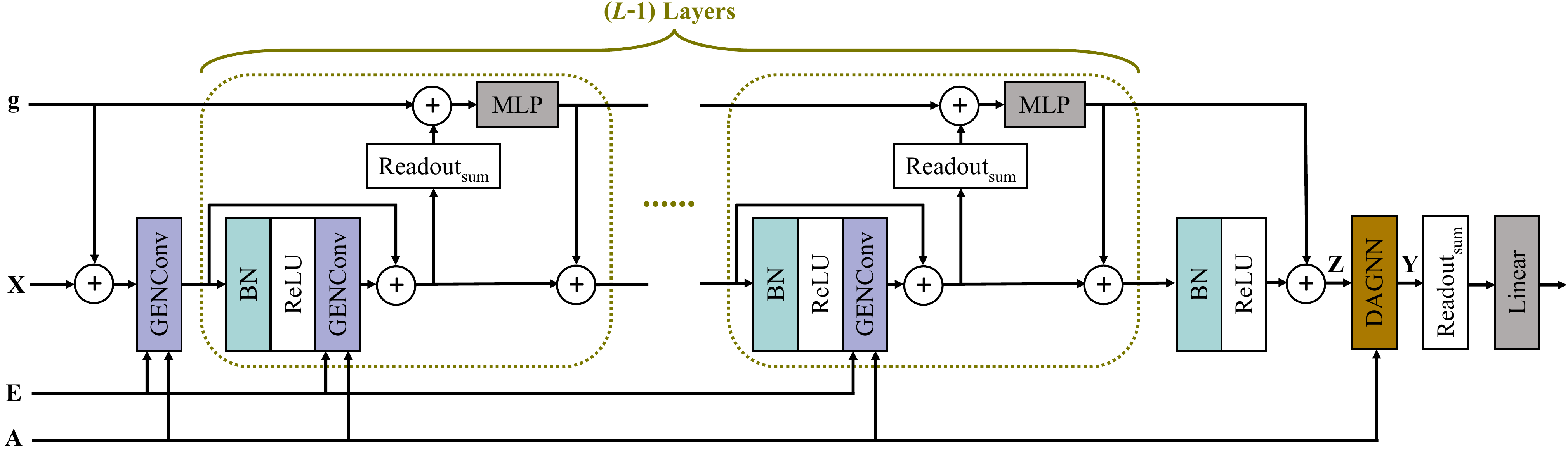}
	\caption{An illustration of our deeper 2D model. Details are described in Section~\ref{sec:2d}.}
	\label{fig:2d_model}
\end{figure*}

Afterwards, we apply a summation readout function to derive a graph-level representation, and then use a linear transformation $\mathbf{W} \in \mathbb{R}^{f \times 1}$ to make prediction for the desired molecular quantum property. Formally, it can be written as
\begin{equation}
    \hat{y} = \textrm{Readout}_\textrm{sum}\left(\{\mathbf{y}_i, i \in \mathcal{V}\}\right)\mathbf{W} \quad \in \mathbb{R}.
\end{equation}
In our 2D model, we have $L$ DeeperGCN Layers and $K$ steps of propagation in DAGNN, thereby making the depth, or receptive field, to be $(L+K)$. The overall architecture of our deeper 2D model is illustrated in Figure~\ref{fig:2d_model}. 

\subsection{3D Model on Low-Cost Conformer Sets}
\label{sec:3d}

Although it is natural to view a molecule as a 2D graph defined by atoms and bonds in chemistry, quantum mechanical simulation methods still require the precise 3D structure of a molecule. 
In particular, the Hamiltonian and the wave functions defining the state of a quantum system are functions of 3D atomic coordinates and are sensitive to small perturbations.
While the 3D equilibrium structure of a molecule, from which we compute the quantum properties, is very expensive to obtain, 3D atomic coordinates that are close to equilibrium can be efficiently generated with open-source software~\citep{ebejer2012freely}.
However, we empirically observe that using only one randomly sampled conformer as input leads to unexpected prediction results.
Consequently, we propose to sample multiple conformers for each molecule and leverage a 3D GNN to learn from the set of conformers. Intuitively, the idea is consistent with the fact that molecules are intrinsically flexible objects with fluctuating 3D structures.
Therefore, learning from a set of conformers can reduce the variance in the input space.

\begin{figure*}[t]
     \centering
          \begin{subfigure}[b]{0.55\textwidth}
         \centering
         \includegraphics[width=\textwidth]{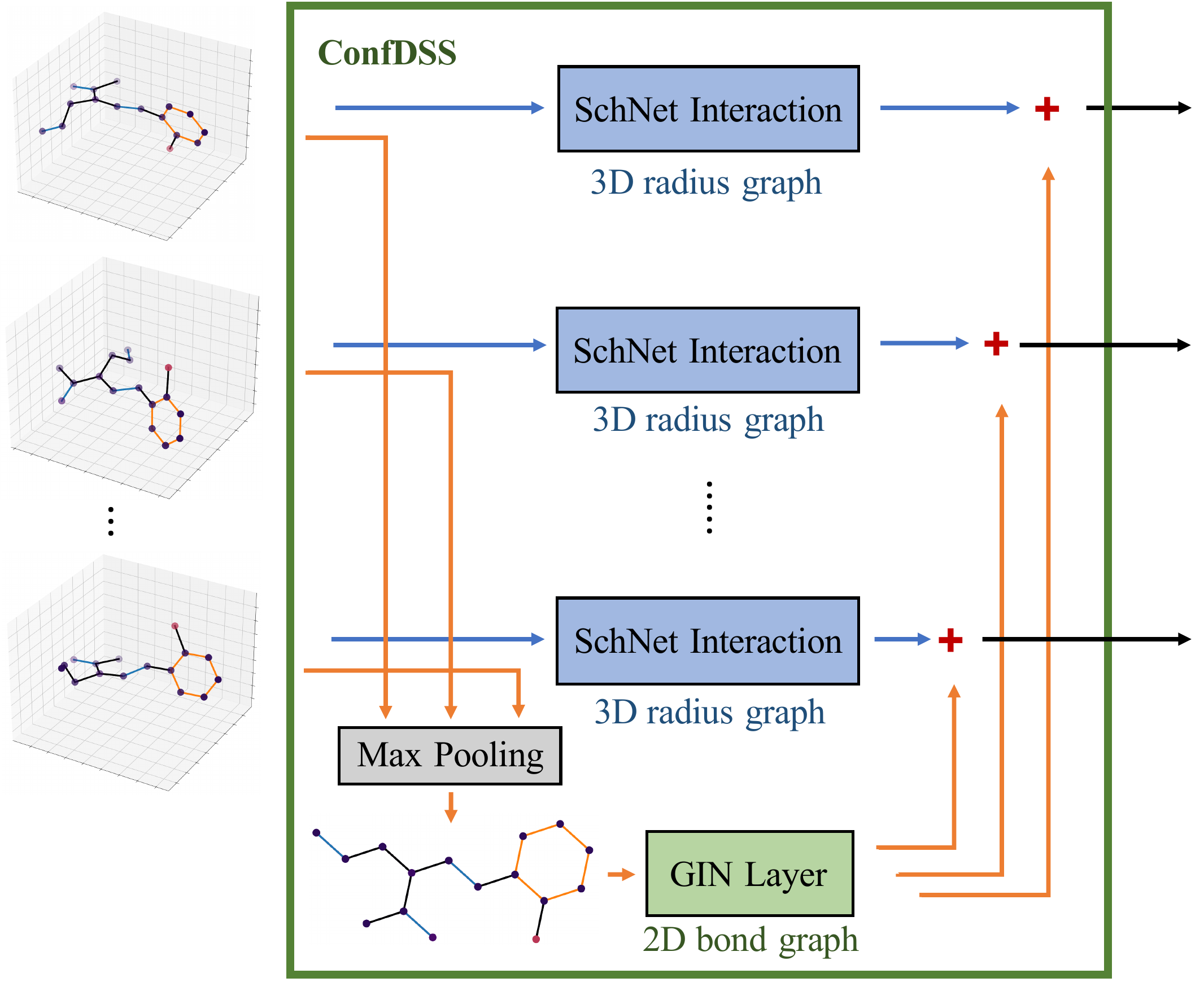}
         \caption{}
         \label{fig:confdss_layer}
     \end{subfigure}\hspace{4em}%
     \begin{subfigure}[b]{0.2\textwidth}
         \centering
        \includegraphics[width=\textwidth]{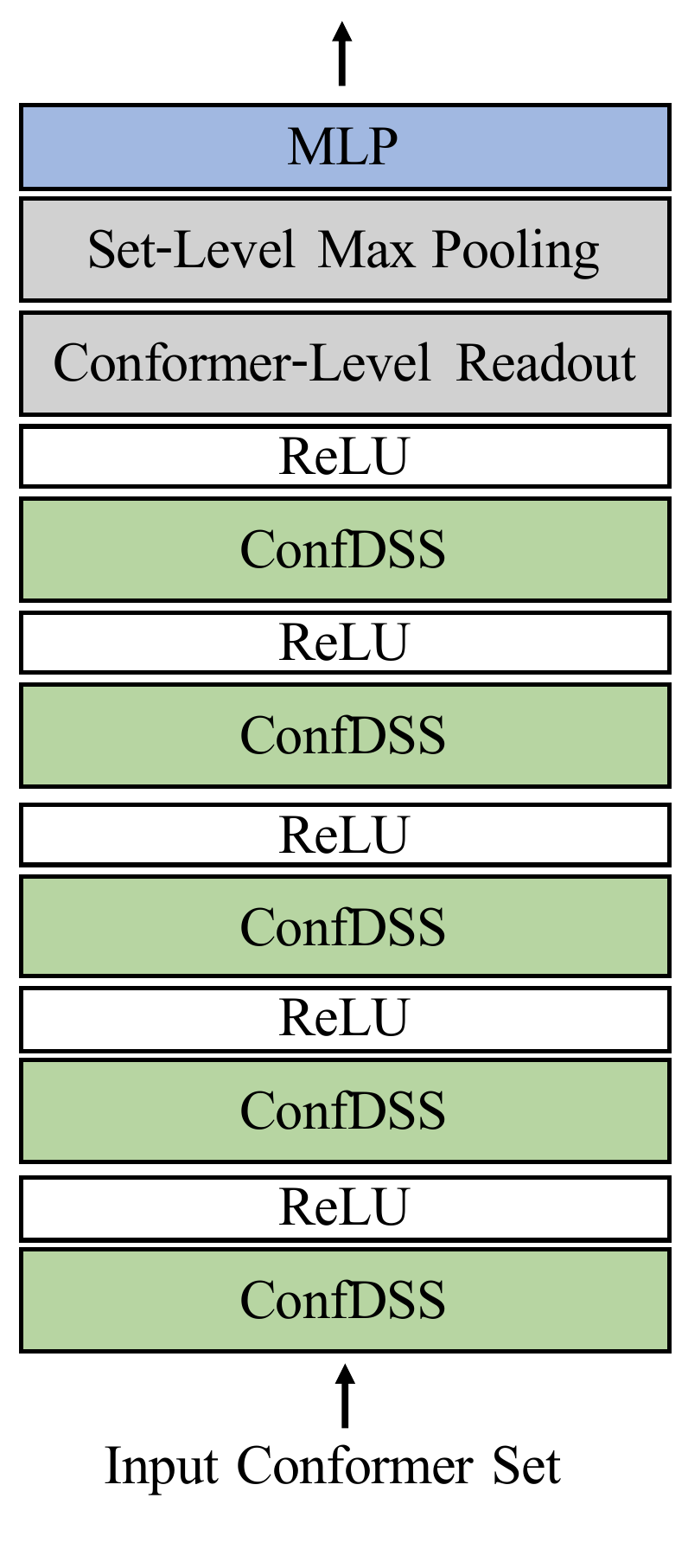}
    	\caption{}
         \label{fig:confdss_net}
     \end{subfigure}
        \caption{Illustrations of (a) the proposed ConfDSS Layer and (b) our 3D model. Details are described in Section~\ref{sec:3d}.}
        \label{fig:3d_model}
\end{figure*}

\paragraph{Conformer Generation.}
We generate mutliple conformers for each molecule with RDKit~\citep{landrum2006rdkit} and prune similar conformers with an RMSD cutoff $R=0.5$\r{A}.

\paragraph{ConfDSS Layer.} Inspired by~\citep{maron2020learning}, we propose the ConfDSS Layer, as illustrated in Figure~\ref{fig:confdss_layer}, to learn from generated conformer sets. Each conformer can be regarded as a set of nodes where each node is associated with a feature vector and a 3D coordinate. Both the input and the output of the ConfDSS Layer are a set of such conformers. Generally, in each ConfDSS Layer, we apply 3D GNN blocks to learn from individual conformers. In addition, as suggested by~\citet{maron2020learning}, a 2D GNN block is deployed to their aggregated graph.

For individual conformers, in order to capture the 3D information, we do not use the graph topology information given by bonds. Instead, we construct a spatial graph where edges are defined between atom pairs closer than a cutoff distance. Individual conformers are then processed by SchNet interaction blocks~\citep{schutt2018schnet}, which can consider the 3D information contained in the conformers. Skip connections are applied after the SchNet blocks. The parameters of the SchNet interaction blocks are shared across conformers. 

Note that the topology information and edge features of the aggregated graph are still defined by bonds. The node features of the aggregated graph are obtained by performing set-level max pooling over the node features of the conformer set. The aggregated graph has the same number of nodes as each conformer. A GIN layer~\citep{xu2018powerful} is then applied to the aggregated graph. Afterwards, we add the node features from the aggregated graph back to the node features of each output conformers to explicitly incorporate the set-level information into each conformer. Similar to our 2D model, we also incorporate the virtual node technique in the ConfDSS layer.

\paragraph{3D Model.} As shown in Figure~\ref{fig:confdss_net}, the final 3D model is composed of 5 ConfDSS Layers follwed by the ReLU activation. For the input of the first ConfDSS layer, all conformers share the same set of node features and become different in latter layers due to their different 3D coordinates. The final output conformer set is aggregated into a single feature vector representing the input molecule by performing conformer-level readout and set-level max pooling. Finally, we apply an MLP to this feature vector for prediction.

\section{Experiments}

\subsection{Setup}

\paragraph{Dataset.} We evaluate our methods on PCQM4M-LSC~\citep{hu2021ogb}, which is included in the 2021 KDD Cup on OGB Large-Scale Challenge. It is collected for graph-level quantum chemical property prediction, specifically, HOMO-LUMO energy gap prediction without given 3D equilibrium structures. It contains over $3.8$M molecules in total.

\paragraph{Implementation details.} We follow OGB~\citep{hu2020open,hu2021ogb} to initialize the node features and edge features for molecular graphs. To be specific, each node is associated with a 9-dimensional feature, including atom type, chirality, \emph{etc.}. Each edge has a 3-dimensional feature, containing bond type, stereochemistry, and conjugation. For our 2D model, we adopt $L=16$ DeeperGCN Layers and set $K=5$ in DAGNN. We set the number of hidden dimensions to be $600$. We apply dropout to the MLP in Eq.~(\ref{eq:virtual}) with $0.25$ dropout rate. We train the model with the Adam optimizer~\citep{kingma2014adam} for $100$ epochs. The initial learning rate is $0.001$ and decays to $25\%$ every $30$ epochs. For our 3D model, we train the model with the Adam optimizer for $60$ epochs. We also use $0.001$ as the initial learning, but it decays to $10\%$ every $40$ epochs. We use up to $20$ conformers for each molecule during training and $40$ conformers per molecule for prediction. The training batch size is set to $256$ for both our 2D and 3D models. Our implementation is available as part of the MoleculeX\footnote{\url{https://github.com/divelab/MoleculeX}} software package under the \textit{BasicProp/kddcup2021} folder.

\paragraph{Baselines.} Following~\citep{hu2021ogb}, we consider GCN~\citep{kipf2016semi}, GIN~\citep{xu2018powerful}, GCN-Virtual, and GIN-Virtual as baselines. The last two models denotes the variants with adding the virtual node.

\paragraph{Data split.} The original training/validation/test split of PCQM4M-LSC is $80\%/10\%/10\%$ and the labels of the test set are not publicly available. For comparing with baselines, we train our model on the given training set and compare the performance on the standard validation set, given that the results are shown to be consistent across the validation set and the hidden test set~\citep{hu2021ogb}.

Notably, we observe that it is useful to include more available data for training. Hence, we consider using the original validation set for training. On the other hand, model selection is also necessary because the performance could fluctuate over different training epochs. Therefore, we randomly divide the original validation set into $5$ sets. Each of them could be used for validation, and the rest are added into the training set. In this case, we can obtain $5$ different $88\%/2\%/10\%$ splits. For improving performance on the hidden test set, we can train multiple models on these $5$ splits and use the ensemble of the predictions of these models on test set. In this case, we can leverage all the available data, including the original training and validation set, for training, while conducting model selection at the same time using the new $2\%$ validation set on the corresponding split. The reason why we divide the original validation into $5$ sets is that the variance brought by the random division is small enough and the numerical result in terms of MAE is similar to the original validation set for a given trained model. For example, a GIN-virtual model trained on the original training set achieves $0.1396$ MAE on the original validation set and $0.1393 \sim 0.1413$ MAE on random $20\%$ of the original validation set.

\subsection{Results}

\begin{wraptable}{R}{9cm}
\vspace{-0.5cm}
  \caption{Results on PCQM4M-LSC in terms of MAE [eV] on the original validation set. The results of baselines are obtained from~\citet{hu2021ogb}. * represents that the MAE is computed on the molecules that have conformers generated by RDKit. There are around $0.1\%$ molecules in the original validation set, for which RDKit cannot generate conformers. The result is still convincing since the proportion of missing data is negligible.}
  \label{tab:result_val}
  \centering
  \begin{tabular}{lcc}
    \toprule
    \textbf{Method} &\textbf{\#Params} &\textbf{Validation} \\
    \midrule
    GCN & $2.0$M & $0.1684$ \\
    GCN-Virtual & $4.9$M & $0.1510$ \\
    GIN & $3.8$M & $0.1536$ \\
    GIN-Virtual & $6.7$M & $0.1396$ \\
    \midrule
     Our 2D model & $34.1$M & $\textbf{0.1278}$ \\
     Our 3D model & $9.0$M & $\textbf{0.1295}^*$ \\
     \midrule
     Our 2D model without DAGNN& $34.1$M & $0.1350$ \\
    \bottomrule
  \end{tabular}
  \vspace{-0.7cm}
\end{wraptable}

\paragraph{Original split.} The results obtained by training on the given training set and then evaluating on the standard validation set are shown in Table~\ref{tab:result_val}. Both our 2D and 3D models outperform previous state-of-the-art methods with obvious margins. These demonstrate the effectiveness of developing deep and large models and incorporating low-cost descriptive 3D information for large-scale quantum property prediction. We also conduct an ablation study to investigate the improvement of including DAGNN. We construct a model by removing DAGNN from our 2D model. As shown in Table~\ref{tab:result_val}, we observe that the performance degrades a lot without DAGNN. This shows the surprising improvement of incorporating DAGNN, considering that there are only $f$, \emph{i.e.}, 600 in our setting, learnable parameters in this DAGNN component. 

\paragraph{New splits.} As discussed above, we obtain $5$ new splits by moving partial validation data to training set. We train multiple models on such $5$ splits and evaluate them on their corresponding validation sets. To be specific, we train $4$ 2D models and $1$ 3D model on each split. The results are summarized in Table~\ref{tab:result_cv}. According the results of individual models, we can observe that the improvement owned to more training data is remarkable, compared to the results of our 2D and 3D model in Table~\ref{tab:result_val}. In addition, we evaluate the ensemble of predictions obtained by multiple 2D and 3D models on validation set for each split. We find that the ensemble improves the performance obviously, and we also observe that the ensemble of different models outperforms the ensemble of several identical models if the number of used models is the same, which indicates that the 2D and 3D models capture different and complementary information for predicting the HUMO-LOMO energy gap.

For predicting the hidden test set, we ensemble the predictions of the multiple models, including $4$ 2D models and $1$ 3D model, on each split. Afterwards, we take the average over the predictions obtained from all $5$ splits. The MAE of our final prediction on the whole hidden test set is $0.1235$, evaluated by the OGB team.

\begin{table}
\caption{Results on PCQM4M-LSC in terms of validation MAE [eV] on the $5$ new splits. As the results in Table~\ref{tab:result_val}, results of our 3D model are still computed on the molecules that have conformers generated by RDKit. If a molecule do not have available conformers, we ignore the 3D model prediction in the ensemble model.}
  \label{tab:result_cv}
  \centering
  \begin{tabular}{lccccc}
    \toprule
    \textbf{Method} &\textbf{1st split} & \textbf{2nd split} & \textbf{3rd split} & \textbf{4th split} & \textbf{5th split}\\
    \midrule
    \multirow{4}{*}{Our 2D model} & $0.1217$ & $0.1181$ & $0.1201$ & $0.1222$ & $0.1192$\\
    & $0.1191$  & $0.1188$ & $0.1186$ & $0.1181$ & $0.1194$\\
    & $0.1176$ & $0.1190$ & $0.1191$ & $0.1188$ & $0.1183$\\
    & $0.1190$ & $0.1185$ & $0.1192$ & $0.1194$ & $0.1198$\\
    \midrule
     Our 3D model & $0.1214$ & $0.1208$ & $0.1216$ & $0.1222$ & $0.1214$ \\
    \midrule
     Ensemble & $0.1117$ & $0.1113$ & $0.1120$ & $0.1111$ & $0.1114$ \\
    \bottomrule
  \end{tabular}
\end{table}

\section{Conclusion}

To predict molecular quantum chemical properties without known 3D equilibrium structures, in this work, we propose a deeper 2D GNN to endow us larger receptive fields and more expressivity on 2D molecular graphs. In addition, to explicitly consider 3D information, we propose a 3D GNN to learn from low-cost conformer sets, which can be obtained in an affordable budget. We perform experiments on the 2021 KDD Cup on OGB Large-Scale Challenge, in which the task is to predict the HOMO-LUMO energy gap of molecules. The results demonstrate that our proposed 2D and 3D methods are effective. We also repartition the available data and adopt ensemble strategy to further improve the prediction performance, which is shown to be helpful according to our experiments.

\section*{Acknowledgements}

This work was supported in part by National Science Foundation grants IIS-2006861, IIS-1955189, IIS-1908198, DBI-1922969, and National Institutes of Health grant 1R21NS102828.

\bibliographystyle{unsrtnat}
\bibliography{my_reference}  






\end{document}